\documentclass[10pt, a4paper]{article}

\usepackage{comment}
\usepackage{float}
\usepackage{bm}
\usepackage[dvipsnames]{xcolor}
\usepackage{amsmath}
\usepackage{amssymb}
\usepackage{url}
\usepackage{booktabs}

\usepackage{todonotes}

\usepackage{lrec-coling2024} 

\usepackage{mathtools}
\usepackage{enumitem}
\usepackage[font=small]{subfig}

\usepackage{cleveref}
\crefname{section}{\S}{\S\S}
\Crefname{section}{\S}{\S\S}
\crefname{table}{Table}{}
\crefname{figure}{Figure}{}
\crefname{algorithm}{Algorithm}{}
\crefname{equation}{}{}
\crefname{appendix}{App.}{}
\crefname{prop}{Proposition}{}
\crefformat{section}{\S#2#1#3}

\usepackage{multirow}

\newcommand{\defn}[1]{{\textbf{#1}}}

\newcommand{\word}[1]{\textit{#1}}
\newcommand{\concept}[1]{\textsc{#1}}

\newcommand{\sense}[2]{\word{#1}\textsubscript{$#2$}}
\newcommand{\synonym}[1]{\textit{#1}}

\newcommand{\calP}{{\mathcal{P}}}
\newcommand{\calS}{\mathcal{S}}

\newcommand{\real}{\mathbb{R}}
\newcommand{\ve}{\mathbf{e}}

\newcommand{\vd}{\mathbf{d}}

\newcommand{\vx}{\mathbf{x}}
\newcommand{\vy}{\mathbf{y}}

\newcommand{\vh}{\mathbf{h}}

\newcommand{\vw}{\mathbf{w}}
\DeclareMathOperator*{\score}{score}
\DeclareMathOperator*{\biaff}{biaff}
\DeclareMathOperator*{\MLP}{MLP}

\DeclareMathOperator*{\ARI}{ARI}

\newcommand{\setsize}[1]{|{#1}|}

\newcommand{\literalcolour}{Brown}
\newcommand{\relatedcolour}{Bittersweet}
\newcommand{\metaphoricalcolour}{Purple}

\newcommand{\literallabel}{Prototype}

\newcommand{\relatededgelabel}{Metonymy}

\newcommand{\metaphoricaledgelabel}{Metaphor}

\newcommand{\tabspace}{\addlinespace[0.6em]}
\newcommand{\slipspace}{\addlinespace[0.1em]}

\usepackage{adjustbox}
\usepackage{array}

\newcolumntype{R}[2]{%
    >{\adjustbox{angle=#1,lap=\width-(#2)}\bgroup}%
    l%
    <{\egroup}%
}

\newcommand{\feature}[1]{\scalebox{.8}[1.0]{\texttt{#1}}}
\newcommand{\yesfeature}[1]{\textcolor{Green}{\feature{#1}}}
\usepackage[normalem]{ulem}
\newcommand{\nofeature}[1]{\textcolor{Red}{\sout{\feature{#1}}}}
\newcommand{\maybefeature}[2]{\textcolor{Green}{\feature{#1}} \textcolor{Orange}{\feature{#2}}}

\tikzset{definition/.style={align=left, shape=rectangle,draw=black, font={\footnotesize}, fill=black!0!white}}
\tikzset{invis/.style={align=left, shape=rectangle,draw=none, font={\footnotesize}, fill=none,text opacity=0}}
\tikzset{lit_definition/.style={anchor=center, align=left, shape=rectangle,draw=\literalcolour, font={\footnotesize}, label={[label distance=-.05cm]:{\footnotesize{\textcolor{\literalcolour}{\literallabel}}}}, fill=black!0!white}}

\tikzset{met_definition/.style={align=left, shape=rectangle,draw=\metaphoricalcolour, font={\footnotesize}, fill=black!0!white}}
\tikzset{ass_definition/.style={align=left, shape=rectangle,draw=\relatedcolour, font={\footnotesize}, fill=black!0!white}}
\tikzset{ghost/.style={ass_definition, dashed}}
\tikzset{ghost_core/.style={lit_definition, dashed}}

\tikzset{split/.style={-, dashed, line width=.3mm, Gray}}
\tikzset{start/.style={->, line width=.3mm, \literalcolour}}
\tikzset{analogy/.style={->, line width=.3mm, \metaphoricalcolour}}
\tikzset{related/.style={->, line width=.3mm, \relatedcolour}}
\tikzset{unknown/.style={->, line width=.3mm}}

\tikzset{derivation/.style={node distance = .7cm and 
1cm}}
\tikzset{tight/.style={node distance = .55cm and .5cm}}
\tikzset{close/.style={node distance = .6cm and 
.3cm}}

\newcommand{\featuregoesto}{\textcolor{\metaphoricalcolour}{$\rightarrow$}}
\everypar{\looseness=-1}

\usepackage{tikz}
\usepackage{tikz-dependency}

\defcitealias{chu1965shortest}{Chu--Liu}

\title{\textit{ChainNet}: Structured Metaphor and Metonymy in WordNet}

\name{Rowan Hall Maudslay$^{1,2,3}$~~Simone Teufel$^1$~~Francis Bond$^4$~~James Pustejovsky$^5$}
\address{$^1$University of Cambridge \; $^2$Magdalene College \; $^3$The Alan Turing Institute  \\
$^4$Palack\'{y} University \; $^5$Brandeis University \\
\texttt{rh635@cam.ac.uk\; sht25@cam.ac.uk\; bond@ieee.org\; jamesp@brandeis.edu}\\}

\abstract{
The senses of a word exhibit rich internal structure. 
In a typical lexicon, this structure is overlooked: a word's senses are encoded as a list without inter-sense relations. 
We present ChainNet, a lexical resource which for the first time explicitly identifies these structures.
ChainNet expresses how senses in the Open English Wordnet are derived from one another: every nominal sense of a word is either connected to another sense by metaphor or metonymy, or is disconnected in the case of homonymy. 
Because WordNet senses are linked to resources which capture information about their meaning, ChainNet represents the first dataset of grounded metaphor and metonymy. 
 \\ \newline  }

\begin{document}

\maketitleabstract

\section{Introduction} \label{sec:intro}

A fundamental feature of language is \defn{polysemy}: in different contexts words exhibit different meanings. 
Consider these invocations of the word \word{march}:
\begin{enumerate}[resume, label={(\arabic*)}]
\itemsep0em 
\item \word{March} always has dreadful rainfall! \label{ex:homonymy}
\item The infantry performed a militant \word{march}.\label{ex:core}
\end{enumerate}
In example~\ref{ex:homonymy} \word{march} refers to a month of the year, while in~\ref{ex:core} \word{march} refers to the act of walking in a regimented way. 
Because these meanings are semantically unrelated, they are said to exhibit \defn{homonymy}. 
By contrast, word meanings can also be extended productively:
\begin{enumerate}[resume, label={(\arabic*)}]
\itemsep0em 
\item The \word{march} approached the town hall. \label{ex:metonymy}
\end{enumerate}
This meaning of \word{march} is closely related to the meaning of \word{march} in \ref{ex:core}, but the semantic type \citep{pustejovsky1995generative} has changed: while in example~\ref{ex:core} \word{march} refers to an act, in \ref{ex:metonymy} it refers to a group of people who are engaged in this act together. 
This kind of meaning change is known as \defn{metonymy}. 
Words can also create new meanings according to other processes, such as \defn{metaphor}. 
A metaphor occurs when, in a creative shift, a word is used in an entirely new setting:
\begin{enumerate}[resume, label={(\arabic*)}]
\itemsep0em 
\item Nothing can stop the \word{march} of science. \label{ex:metaphor}
\end{enumerate}
In~\ref{ex:metaphor}, the imagery of \ref{ex:metonymy} is used metaphorically to describe the steady progress of an academic discipline. 
Taken together, these examples illustrate how word meaning can be successively extended: \ref{ex:core} was extended to \ref{ex:metonymy}, which in turn was extended to \ref{ex:metaphor}.
This is called \defn{chaining} \citep{lakoff1987women}. 
Chaining can cause word meaning to extend into disjointed areas of semantic space \citep[p.\ 72]{austin1961meaning}: the academic progress in \ref{ex:metaphor} and the regimented walking in \ref{ex:core} have little in common, even though they are connected in a chain.

\begin{figure}
\centering
\begin{tikzpicture}[derivation]
\node[lit_definition, text width=2.7cm] (1) at (0,0) {\sense{march}{{2}} [\synonym{marching}] the act of marching; walking with regular steps};
\node[ass_definition, text width=2.cm] (3) [right = 1.6cm of 1] {\sense{march}{{4}} a procession of people walking together};
\node[lit_definition, text width=2.9cm] (4) [below right = .6cm and -2.8cm of 1] {\sense{march}{{1}} the month following February and preceding April};

    \path[related] (1) edge node[anchor=south, font={\footnotesize}] {\relatededgelabel} (3);
\node[met_definition, text width=2.6cm] (2) [below = .65cm of 3] {\sense{march}{{3}} a steady advance, e.g.\ ``the march of science''};
        \path[analogy] (3) edge node[anchor=west, font={\footnotesize}] {\metaphoricaledgelabel} (2);
\end{tikzpicture}
\caption{The first four senses of \word{march} in ChainNet}
\label{fig:march}
\end{figure}
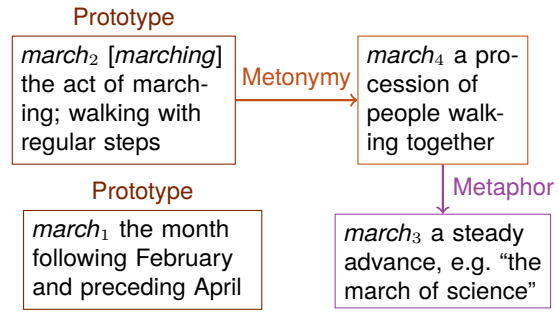

We present \defn{ChainNet}, a new resource of English metaphor, metonymy, and homonymy. 
To formalise chaining, we use a lexicon.
In a typical lexicon a word is represented by a quasi-ordered list of senses which it can be used to evoke. 
For example, below are the first four nominal senses of \word{march} from WordNet 3.0 \citeplanguageresource{_Fellbaum:1998}, each corresponding to one of the previous examples: 
\noindent\begin{footnotesize}
\begin{tabular}{l@{~~~}p{6.1cm}} \tabspace
\sense{march}{{1}} & the month following February 
\\ \tabspace 
\sense{march}{{2}} & [\synonym{marching}] the act of marching; walking with regular steps (especially in a procession of some kind), e.g.\ ``it was a long march'' \\ \tabspace 
\sense{march}{{3}} & a steady advance, e.g.\ ``the march of science'', ``the march of time''
\\ \tabspace
\sense{march}{{4}} & a procession of people walking together, e.g.\ ``the march went up Fifth Avenue'' \\ \tabspace
\end{tabular}
\end{footnotesize}
In ChainNet, the underlying structure of these senses is revealed (\cref{fig:march}): every sense is either a prototypical sense, or it is derived from another sense by metaphor or metonymy.

Unlike other metaphor resources, ChainNet provides a representation of the meaning change that is caused by each metaphor.
This takes the form of a \defn{feature transformation}.
More specifically, features are associated with the sense that a metaphor extends, and the transformation from this sense to a metaphor can then be traced by parallel changes in features.
For the ``\word{march} of science'' metaphor (\sense{march}{3}), the feature transformation is:
\begin{center} \vspace{-1em}

    \begin{tabular}{l@{~~}c@{~~}l} \slipspace
    \textbf{Features of \sense{march}{4}}  &  & \textbf{Features of \sense{march}{3}} \\ 
        \feature{gradually advances} & \featuregoesto &  \yesfeature{gradually advances} \\ 
       \feature{is a group of people} & \featuregoesto &  \nofeature{is a group of people} \\ \slipspace
    \end{tabular} 
\end{center}
While the first feature of \sense{march}{4} is maintained, the second feature is lost by the metaphorical extension.
This information can be used in combination with other multimodal resources already linked to WordNet (including images) to capture the meaning transfer of each metaphor in ChainNet.

We produce ChainNet annotations for every nominal sense of $6500$ words in WordNet, for a total of $22{,}178$ senses.
To collect this annotation, we employed three annotators, who we also used for an agreement study (\cref{sec:annotation}).
We establish two baselines for the task of synthesising ChainNet annotations (\cref{sec:baseline}). 
Finally, we discuss the lessons learned from applying the theory of chaining \citep{lakoff1987women} to real-world data (\cref{sec:discussion}).

\section{Existing Resources}

We are not aware of any computational resource built for research on chaining. 
Most chaining theories are supported only by toy examples \citep{wittgenstein1953philosophical, austin1961meaning, lakoff1987women}, and to the best of our knowledge no chaining formalism has been applied to large amounts of real-world data.  

Metaphor resources are more common. 
Most consist of sentences in which words are labelled for metaphoricity.
The best-known example is the VUAMC \citeplanguageresource{steen2010metaphor}, which was annotated based on the Metaphor Identification Procedure \citep[MIP;][]{semino2007mip}. 
The VUAMC has been used in two shared tasks to build models that identify metaphorical words \citep{leong-etal-2018-report,leong-etal-2020-report}.
However, the main drawback of resources like the VUAMC is that they do not provide any representation of metaphor meaning. 
As a result, metaphor understanding is understudied. 

Instead of analysing tokens, an alternative approach is to analyse senses in a lexicon.
It is impossible for a lexicon to enumerate all possible metaphors \citep{black1962metaphor, black1977more}, but many lexica do nevertheless contain examples of \defn{conventional metaphor}. 
These are metaphors that have been widely adopted by a language community. 

\defn{WordNet} \citeplanguageresource{miller1995wordnet} is a popular computational lexicon. 
WordNet contains senses that correspond to conventional metaphors (e.g.\ \sense{march}{3} in \cref{sec:intro}), but does not systematically identify which senses these are. 
Sometimes a sense's gloss mentions that it is metaphorical, but the choice to include this information was left to the discretion of the lexicographers. 
Consequently, most metaphors are unlabelled, and metonymy is not labelled at all.

WordNet is well positioned for research on chaining and metaphor understanding.
This is because WordNet senses are linked to multimodal data that captures their meaning, in the form of images \citep[e.g.\ ImageNet,][]{deng2009imagenet} and textual usages \citep[see][]{petrolito-bond-2014-survey}. 
WordNet could also be used for \defn{conceptual metaphor theory} \citep[CMT;][]{lakoff1980metaphors} research, as has been suggested by \citet{alonge2002way} and \citet{lonneker2003way}.
A conceptual metaphor is a metaphor pattern that affects multiple words (e.g.\ \concept{argument is war}, affecting words like \word{attack} and \word{defend}).
Currently the main resource used for CMT research is the Master Metaphor List \citep{lakoff1994master}, but this list is incomplete and is not designed for computational work.
An ongoing effort to address this is MetaNet \citep{petruck2018metanet} but this is not openly available.
The benefits of using WordNet to study conceptual metaphor could also apply to \defn{logical metonymy}, which is the equivalent term used to describe metonymy patterns \citep{pustejovsky1995generative}.

Some sections of WordNet have been annotated for metaphoricity, but only a few word types are covered: $440$ in \citet{mohammad-etal-2016-metaphor} and $198$ in \citet{maudslay-teufel-2022-metaphorical}.
We are not aware of any resource (based on WordNet or otherwise) which systematically identifies metaphor or metonymy through the lexis, and which is also computationally tractable and open. 
An exception is the Pattern Dictionary of English Verbs \citep{pustejovsky-etal-2004-automated, hanks2005pattern, hanks2013lexical}, which identifies semantic type patterns of English verbs but is not connected to WordNet.

\section{The ChainNet Schema} \label{sec:chainnet}

ChainNet is a new resource that formalises chaining.
It is built as an annotation layer for the Open English Wordnet \citep{mccrae-etal-2019-english}\nocitelanguageresource{oewn:2020}, itself an extension to Princeton WordNet 3.0 \citeplanguageresource{_Fellbaum:1998}. 

\subsection{Three Types of Sense} \label{sec:labels}

In ChainNet, every nominal sense of a word is either a prototypical sense or is derived from another sense. 
We permit only two derivation types: metaphor and metonymy.
The idea that metaphor and metonymy are the two axes of meaning extension goes back at least to \citet{jakobson1956two}, who saw metaphor as a ``vertical'' relation based on selection, substitution, and similarity, and saw metonymy as a ``horizontal'' relation based on combination, contexture, and contiguity. 
Various authors have proposed further subdivisions to this classification, adding categories for synecdoche \citep[e.g.][]{bloomfield1933language, blank1999typology}, folk-etymology \cite{ullman1962semantics, blank1999typology}, specialisation and generalisation \citep{bloomfield1933language, blank1988metaphors}, and so on. 
For the first iteration of ChainNet, however, we choose to empirically test the descriptive power of a simple metaphor--metonymy dichotomy.

The output for each word is therefore a rooted tree (where the prototypical sense is the root), or multiple disjoint trees in the case of homonymous words (as these have multiple roots). 
A partial example for \word{march} was shown in \cref{fig:march}; in the spirit of \citeauthor{jakobson1956two}, we depict metaphor as a vertical relation and metonymy as a horizontal relation.\footnote{Any words in square brackets at the start of a definition correspond to the synonyms in WordNet synsets.}

\begin{figure*}[htb]
\small
    \centering
    \begin{tikzpicture}[derivation]
\node[invis, text width=5.5cm] (3) at (0,0) {\sense{neck}{{1}} [\synonym{cervix}] the part of an organism (human or animal) that connects the head to the rest of the body, e.g.\ ``he admired her long graceful neck''};
\node[met_definition, text width=3cm] (5) [below left =.65cm and -2cm of 3] {\sense{neck}{{2}} a narrow elongated projecting strip of land};
\node[invis, text width=2.5cm] (5_above) [above =.65cm of 5] {};
\path[analogy] (5_above) edge node[anchor=west, font={\footnotesize}] {\metaphoricaledgelabel} (5);
\node[met_definition, text width=5.5cm] (2) [below right=.65cm and -3.3cm of 3] {\sense{neck}{{4}} a narrow part of an artifact that resembles a neck in position or form, e.g.\ ``the banjo had a long neck''};
\node[invis, text width=2.5cm] (2_above) [above =.65cm of 2] {};
\path[analogy] (2_above) edge node[anchor=west, font={\footnotesize}] {\metaphoricaledgelabel} (2);
\node[ass_definition, text width=1.9cm] (4) [left = 1.6cm of 3] {\sense{neck}{{3}} a cut of meat from the neck of an animal};
    \path[related] (3) edge node[anchor=south, font={\footnotesize}] {\relatededgelabel} (4);\node[ass_definition, text width=4.5cm] (1) [right = 1.6cm of 3] {\sense{neck}{{5}} [\synonym{neck opening}] an opening in a garment for the neck of the wearer; a part of the garment near the wearer's neck};
    \path[related] (3) edge node[anchor=south, font={\footnotesize}] {\relatededgelabel} (1);

\node[lit_definition, text width=5.5cm] (3_render) at (0,0) {\sense{neck}{{1}} [\synonym{cervix}] the part of an organism (human or animal) that connects the head to the rest of the body, e.g.\ ``he admired her long graceful neck''};
\end{tikzpicture}
    \caption{The complete ChainNet structure of \word{neck}}
    \label{fig:neck}
\end{figure*}
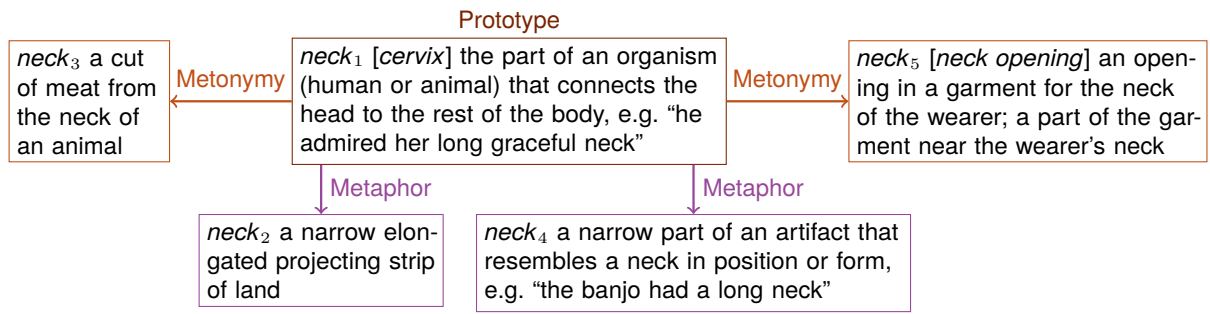

\subsubsection{\textcolor{\literalcolour}{Prototypes}} \label{sec:core} 

In ChainNet every word has a prototypical sense, which is a central sense from which other senses expand.\footnote{We adopt the term ``prototypical sense'' following \citet{lakoff1987women}, who himself adapted it from \citet{rosch1973natural}.
Other terms that are sometimes used include ``nuclear sense'' \citep{austin1961meaning}, ``core sense'' \citep[e.g.][]{pedersen-etal-2022-compiling}, ``sanctioning sense'' \citep[e.g.][]{evans2005meaning}, or ``basic sense'' \citep[e.g.][]{copestake1995semi}.}
Following MIP, we define a prototypical sense as a ``basic'' sense of a word.
For a group of related senses, only one is prototypical. To determine which, we create criteria that annotators can apply. A prototypical sense is usually:

\begin{enumerate}
\item \textbf{Evoked out of context.}
A prototypical sense is usually what comes to mind when a word is heard on its own, without additional context. 

\item \textbf{Referenced by the unigram.} 
A prototypical sense is usually referenced using a word on its own, rather than in a multi-word phrase. 
For example, \word{apple} can refer to either a fruit (\sense{apple}{1}) or a tree which bears that fruit (\sense{apple}{2}).
To refer to the fruit, speakers would likely use the word alone, and say e.g.\ ``I ate the \word{apple}'' rather than ``I ate the \word{apple fruit}''. The inverse is true for the tree: ``I watered the \word{apple tree}'' is more likely than ``I watered the \word{apple}''.
Because of this, the fruit (\sense{apple}{1}) is prototypical.

\item \textbf{Related to the physical world.} 
A prototypical sense is usually concrete or imageable. 

\item \textbf{Related to human experience.} 
A prototypical sense is often related to embodied life.

\item \textbf{Historically older than other senses.} 
A prototypical sense usually predates other senses in terms of their etymology.

\end{enumerate}
A word can have multiple prototypical senses, but this only occurs when the word exhibits homonymy.
An example is \word{bridge}, which refers to a crossing over a river (\sense{bridge}{1}) and a card game (\sense{bridge}{5}):

\noindent \begin{footnotesize}
\begin{tabular}{l@{~~~}p{6cm}} \tabspace
\sense{bridge}{1} & [\synonym{span}] a structure that allows people or vehicles to cross an obstacle such as a river 
\\ \tabspace
\sense{bridge}{5} & any of various card games based on whist 
\\ 
\tabspace
\end{tabular}
\end{footnotesize}
These senses have different etymological origins and are unrelated, so they are both prototypical.

Our five criteria resemble other guidelines (e.g.\ MIP or \citealp{evans2005meaning}, p.\ 44).
However, a notable point of difference is that we tell annotators explicitly to prioritise a sense which is not historically the oldest (i.e.\ to violate criteria 5) if this would be the prototypical sense in the minds of contemporary speakers. 
For example, \word{train} can refer to a locomotive (\sense{train}{1}) or a procession of camels (\sense{train}{3}):

\noindent \begin{footnotesize}
\begin{tabular}{l@{~~~}p{6.2cm}}\tabspace
\sense{train}{1} & [\synonym{railroad train}] public transport provided by a line of railway cars coupled together and drawn by a locomotive, e.g.\ ``express trains don't stop at Princeton Junction''  
\\ \tabspace
\sense{train}{3} & [\synonym{caravan}, \synonym{wagon train}] a procession (of wagons or mules or camels) traveling together in single file, e.g.\ ``they joined the wagon train for safety'' 
\\ \tabspace
\end{tabular}
\end{footnotesize}
The second sense predates the first, but most people today would think of the locomotive, so it is the prototypical sense. 
Similar logic applies when determining whether two senses are homonymous. 
Consider a third sense of \word{bridge}:

\noindent \begin{footnotesize}
\begin{tabular}{l@{~~~}p{6cm}} \tabspace
\sense{bridge}{9} & [\synonym{bridge deck}] an upper deck where a ship is steered and the captain stands  \\ \tabspace
\end{tabular}
\end{footnotesize}
This sense originated on paddle steamers which had a bridge across the deck: it is derived from \sense{bridge}{1}. 
However, today most speakers would not perceive a relation between \sense{bridge}{9} and \sense{bridge}{1}, so \sense{bridge}{9} is also a prototypical sense.
In this way, ChainNet focuses explicitly on cognitive versions of linguistic phenomena rather than etymological versions, while other resources conflate the two. 

\subsubsection{\textcolor{\metaphoricalcolour}{Metaphors}} \label{sec:metaphorical}

Metaphor occurs when one sense resembles another, often in an abstract way. 
Consider \word{neck}:

\noindent \begin{footnotesize}
\begin{tabular}[b]{l@{~~~}p{6.2cm}} \tabspace
\sense{neck}{1} & [\synonym{cervix}] the part of an organism (human or animal) that connects the head to the rest of the body, e.g.\ ``he admired her long graceful neck'', ``the horse won by a neck'' \\ \tabspace
\sense{neck}{2} & a narrow elongated projecting strip of land  \\ \tabspace
\sense{neck}{4} & a narrow part of an artifact that resembles a neck in position or form, e.g.\ ``the banjo had a long neck'', ``the bottle had a wide neck''  \\ \tabspace
\end{tabular}
\end{footnotesize}
\noindent The prototypical sense (\sense{neck}{1}) refers to the part of the body connecting the head to the torso. 
The other senses \sense{neck}{2} (a long strip of land) and \sense{neck}{4} (a narrow part of object) are both ``like'' \sense{neck}{1} (they are long and thin), and so they are connected by metaphor.
This is shown in \cref{fig:neck}.

We operationalise metaphor as a relation between a pair of senses, in order to capture how metaphor connects one meaning to another. 
This is a divergence from existing resources, which assign a binary label (either metaphorical or literal) to individual senses \citep[e.g.][]{mohammad-etal-2016-metaphor, maudslay-teufel-2022-metaphorical} or to individual tokens (e.g.\ the VUAMC). 
This approach to annotation often creates problems, because it does not account for chaining.
For example, if sense C was a metaphorical extension of B, and B was itself a metaphorical extension of A, then B would be both ``metaphorical'' (compared to A) \underline{and} ``literal'' (compared to C). 
These cases can be represented in ChainNet, which has no notion of literal senses.

\subsubsection{\textcolor{\relatedcolour}{Metonymies}} \label{sec:associated}

Metonymy occurs when a sense is related to another, but has a different semantic type \citep{pustejovsky1995generative}.
Consider two more senses of \word{neck}:

\noindent \begin{footnotesize}
\begin{tabular}{l@{~~~}p{6.2cm}} \tabspace
\sense{neck}{3} & a cut of meat from the neck of an animal \\ \tabspace
\sense{neck}{5} & [\synonym{neck opening}] an opening in a garment for the neck of the wearer; a part of the garment near the wearer's neck \\ \tabspace
\end{tabular}
\end{footnotesize}
\noindent These senses are not ``like'' \sense{neck}{1}, but they are related to it in other ways. The sense \sense{neck}{3} is ``meat from'' \sense{neck}{1}, and \sense{neck}{5} is ``a hole in clothing for'' \sense{neck}{1}: they are metonyms (see \cref{fig:neck}).

The full description of metonymy that we give to annotators is broad. It encompasses logical metonymy schemas \citep{pustejovsky1995generative}, such as count--mass alternations (\sense{neck}{1} to \sense{neck}{3}) or figure--ground reversals (\sense{neck}{1} to \sense{neck}{5}). It also encompasses synecdoche, which is when the word for part of something is used to refer to the whole of it, for example when \word{wheels} refers to a whole vehicle:
\begin{center}
\begin{tikzpicture} [derivation] 
    \node[lit_definition, text width=3.9cm] (a) at (0,0) {\sense{wheel}{1} a simple machine consisting of a circular frame with spokes (or a solid disc) that can rotate on a shaft or axle (as in vehicles or other machines)};
    \node[ass_definition, text width=2.9cm] (b) [right = .3cm of a] {\sense{wheel}{7} [\synonym{bicycle}, \synonym{bike}, \synonym{cycle}] a wheeled vehicle that has two wheels and is moved by foot pedals};  
    \path[related] (a) edge node[anchor=south, font={\footnotesize}] {} (b); 
\end{tikzpicture} 
\end{center}

\subsection{Recording Metaphor Meanings} \label{sec:features}

ChainNet provides an explanation of how each metaphor modifies the meaning of the sense it extends.
The way that explanations are encoded is based on \defn{slippage} \citep{hofstadter1995fluid}. 
Consider \word{leaf}, which refers to a plant organ (\sense{leaf}{1}) or a page in a book (\sense{leaf}{2}):
\begin{center}
{\begin{tikzpicture} [derivation] 
    \node[lit_definition, text width=7cm] (a) at (0,0) {\sense{leaf}{1} [\synonym{leafage}, \synonym{foliage}] the main organ of photosynthesis and transpiration in higher plants};
    \node[met_definition, text width=7cm] (b) [below = .55cm of a] {\sense{leaf}{2} [\synonym{folio}] a sheet of any written or printed material (especially in a manuscript or book)};  
        \path[analogy] (a) edge node[anchor=west, font={\footnotesize}] {\metaphoricaledgelabel} (b);

\end{tikzpicture}}
\end{center}
The metaphorical extension \sense{leaf}{2} has some features in common with \sense{leaf}{1} (e.g.\ they are both flat), but it is missing an important feature (it is not part of a plant): this feature has ``slipped''.
For every metaphor, ChainNet records features that are maintained, and features that slip.

\subsubsection{Procedure}
Suppose we have a pair of senses linked by metaphor, which for convenience we refer to here as the literal sense (e.g.\ \sense{leaf}{1}) and the metaphorical sense (\sense{leaf}{2}).
Annotators first add features to the literal sense. 
These features are not intended to be a set of necessary and sufficient conditions, but instead are meant to capture specific nuances of meaning that are relevant to the metaphorical transformation. 
Features are given in natural language, and are elicited from the annotator with a sentence fragment: \feature{This thing \_\_\_}. 
In the case of \word{leaf}, features of the literal sense might be \feature{is flat}, \feature{photosynthesises}, and \feature{is part of a plant}. 

For the metaphorical sense, annotators then decide whether each of those features is kept, lost, or modified.
For the \word{leaf} example, a complete feature transformation (with one kept feature, one lost feature, and one modified feature) could be:
\begin{center} 
    \begin{tabular}{lcl} \slipspace
    \textbf{Features of \sense{leaf}{1}}  &  & \textbf{Features of \sense{leaf}{2}} \\ 
        \feature{is flat} & \featuregoesto &  \yesfeature{is flat} \\ 
       \feature{photosynthesises} & \featuregoesto &  \nofeature{photosynthesises} \\
        \feature{is part of a plant} & \featuregoesto &  \maybefeature{is part of a}{book} \\ \slipspace
    \end{tabular}
\end{center}
At minimum, every metaphor has one modified feature, or one kept feature plus one lost feature.

\subsubsection{Multiple Metaphorical Extensions}

When multiple metaphors extend the same sense, they each transform the same set of features. Consider again \word{train}, which can refer metaphorically to a series of events (\sense{train}{4}) or a bride's gown (\sense{train}{5}):
\begin{center}\vspace{-1.2em}
    \begin{tikzpicture}[derivation] 
        \node[definition, text width=7cm, draw=none] (a) at (0,0) {\sense{train}{1} [\synonym{railroad train}] public transport provided by a line of railway cars coupled together and drawn by a locomotive, e.g.\ ``express trains don't stop at Princeton Junction''};
        \node[met_definition, text width=2.5cm] (b) [below left= .7cm and -2.66cm of a] {\sense{train}{4} a series of consequences wrought by an event, e.g.\ ``it led to a train of disasters''};
        \node[met_definition, text width=4.3cm] (d) [right = .2cm of b] {\sense{train}{5} a piece of cloth forming the long back section of a gown that is drawn along the floor, e.g.\ ``the bride's train was carried by her two young nephews''};
                \node[invis, text width=4.3cm] (d_above) [above = .68cm of d] {\sense{train}{5} a piece of cloth forming the long back section of a gown that is drawn along the floor, e.g.\ ``the bride's train was carried by her two young nephews''};
        \path[analogy] (d_above) edge node[anchor=west, font={\footnotesize}] {\metaphoricaledgelabel} (d); 
                \node[invis, text width=2.5cm] (b_above) [above = .7cm of b] {};

                \draw[analogy] (b_above) edge node[anchor=west, font={\footnotesize}] {\metaphoricaledgelabel} (b); 

                \node[lit_definition, text width=7cm] (a) at (0,0) {\sense{train}{1} [\synonym{railroad train}] public transport provided by a line of railway cars coupled together and drawn by a locomotive, e.g.\ ``express trains don't stop at Princeton Junction''};

    \end{tikzpicture} 
\end{center}
Suppose an annotator has given \sense{train}{1} the features:
\begin{center}
    \begin{tabular}{l} 
        \feature{is a mode of transport}  \\ 
       \feature{is a series of carriages} \\
        \feature{is pulled from the front} \\ 
    \end{tabular}
\end{center}
Each metaphor transforms these features differently. The first metaphor, \sense{train}{4} (consequences), builds on the sequentiality of \sense{train}{1}:
\begin{center}
    \begin{tabular}{cl} 
      \featuregoesto &  \nofeature{is a mode of transport}  \\ 
      \featuregoesto & \maybefeature{is a series of}{consequences} \\
      \featuregoesto &  \nofeature{is pulled from the front} \\ 
    \end{tabular}
\end{center}
The second metaphor, however, is explained by a very different slippage. A \sense{train}{5} (gown) is pulled from the front by the bride, similar to how a \sense{train}{1} is drawn from the front by an engine carriage:
\begin{center}
    \begin{tabular}{cl} 
      \featuregoesto &  \nofeature{is a mode of transport}  \\ 
      \featuregoesto & \nofeature{is a series of carriages} \\
      \featuregoesto &  \yesfeature{is pulled from the front} \\ 
    \end{tabular}
\end{center}
In this way, slippages distinguish between the different types of metaphor that act on the same word. 

\subsection{Making Alterations to the Lexicon}

In rare cases, errors or omissions in WordNet make it difficult to apply our annotation schema. 
In these situations, we allow annotators to modify the senses of a word using two operations. 

\paragraph{Splitting Senses} 
Sometimes, a sense conflates a metaphorical and a non-metaphorical sense: 

\noindent \begin{footnotesize}
\hspace{-.5em}\begin{tabular}{l@{~~}p{6.5cm}} \tabspace
\sense{birth}{1} & the time when something begins (especially life), e.g.\ ``they divorced after the birth of the child'', ``his election signaled the birth of a new age'' 
\\ \tabspace
\end{tabular}
\end{footnotesize}
This sense covers both a metaphorical sense of \word{birth} (``the \word{birth} of a new age''), and a literal sense (``the \word{birth} of the child'').
To fix this, annotators can split the sense into two separate senses: a non-metaphorical sense \sense{birth}{1\text{A}}, and a metaphorical sense \sense{birth}{1\text{B}}. They then edit the definition in each case to make the distinction clear, and annotate each half in isolation.

\paragraph{Virtual Senses} 
Sometimes a chain exists but cannot be expressed because an intermediate sense is missing. 
Consider the word \word{twin}:

\noindent \begin{footnotesize}
\begin{tabular}{l@{~~~}p{6.4cm}}\tabspace
\sense{twin}{{1}} & either of two offspring born at the same time from the same pregnancy
\\ \tabspace
\sense{twin}{{2}} & [\synonym{Gemini}] (astrology) a person who is born while the sun is in Gemini 
\\ \tabspace
\end{tabular}
\end{footnotesize}
The second sense relates to the fact that the Gemini star sign is represented by the twins, Castor and Pollux. 
However, in WordNet \word{twin} does not have a Gemini star sign sense.
To encode the chain, annotators have the option to add an additional virtual sense, which is a new sense, with a definition provided by the annotator:

\begin{center}
\begin{tikzpicture}[derivation] 
    \node[lit_definition, text width=2.2cm] (a) at (0,0) {\sense{twin}{1}  either of two offspring born at the same time from the same pregnancy};
    \node[ghost, text width=2cm] (g) [right = .3cm of a] {\sense{twin}{\text{V}1} [\synonym{Gemini}] a star sign represented by the twins, Castor and Pollux};
    \node[ass_definition, text width=2.1cm] (b) [right = .3cm of g] {\sense{twin}{2}  [\synonym{Gemini}] (astrology) a person who is born while the sun is in Gemini};
        \path[related] (a) edge node[anchor=south, font={\footnotesize}] { } (g);
        \path[related] (g) edge node[anchor=south, font={\footnotesize}] { } (b);

\end{tikzpicture}
\end{center}

\section{Collecting Annotations} \label{sec:annotation}

\subsection{Guidelines and Interface} \label{sec:guidelines}

To collect ChainNet annotations, we produced in-depth annotation guidelines and developed a web-based graphical user interface.\footnote{In resources given to annotators, we adapt some of the nomenclature. Metonyms are referred to as ``associations'', because we found in preliminary work that the similarity in form of ``metaphor'' and ``metonymy'' increased annotation errors. Prototypical senses are referred to as ``core senses'', in order to avoid specialist terminology.}
By default, the interface restricts the types of chain that it is possible to create: metonyms can only extend prototypical senses, and metaphors cannot extend other metaphors. 
However, this constraint can be overridden by marking a sense as a ``conduit'', which allows any sense to connect to it. 
We added this soft constraint as we found in preliminary work that it biased annotators towards shallower trees, which improved agreement.
Further details are in \cref{app:interface}.

\subsection{Data}

To select words for annotation, we first filtered WordNet to only include words with $2$--$10$ nominal senses. 
Monosemous words have only one solution (a single prototypical sense), while words with greater than $10$ senses were deemed to require too great a cognitive load. This is because the number of possible annotation options increases exponentially with the number of senses, as shown in the middle column of \cref{tab:annotationcollected}.
We also excluded single-letter wordforms, wordforms which included whitespace or hyphens, and words whose senses were all proper nouns. 
We then sampled words from the remaining set according to their frequency as nouns, which we estimated by POS-tagging Wikipedia.

\subsection{Procedure}

We employed three annotators, all undergraduate linguists who are native speakers of British English. 
They were selected on the basis of a screening task. 
Annotation consisted of three main phases: training ($100$ words each), bulk 1 ($500$ each), and bulk 2 ($1000$ each). After this, annotators bulk-annotated a further $1900$ words between them.

To evaluate the annotation quality, we measure inter- and intra-annotator agreement.
Inter-annotator agreement evaluates the degree to which different annotators produce the same annotation.
To measure inter-annotator agreement, $100$ shared words were distributed throughout the three main phases. 
Intra-annotator agreement, on the other hand, measures the consistency of annotation within a single mind. 
Intra-annotator agreement can be measured by asking an annotator to re-annotate material they have annotated already, after a suitable time period: at the end of the main phases, we gave each annotator a set of $100$ words which they had annotated $6$~weeks prior.
It stands to reason that tasks that cannot be held consistent within a single mind are even harder to reliably convey to others, so typically intra-annotator agreement is higher than inter-annotator agreement.

The end result is a dataset with $6500$ annotated words, of which $100$ are triply annotated, and $300$ are doubly annotated by one annotator.
WordNet coverage is shown in \cref{tab:annotationcollected}: ChainNet covers a high percentage of words with many senses ($94\%$ of words with $5$--$10$ senses), but for the long tail of low-frequency words with few senses, coverage decreases. 

\begin{table}
\small
    \centering
    \begin{tabular}{c r@{}l r@{ }c@{ }l@{ }l} \toprule
\# senses & \multicolumn{2}{c}{\# options} & \multicolumn{4}{c}{\# words annotated} \\ \midrule
\hphantom{$0$}$2$ & $5$ &  & $2634$ & / & $10259$ & ($26\%$) \\
\hphantom{$0$}$3$ & $49$ & & $1502$ & / & $2988$ & ($50\%$) \\
\hphantom{$0$}$4$ & $729$ &  & $1052$ & / & $1178$ & ($89\%$) \\
\hphantom{$0$}$5$ & $146$ & $\times 10^{2}$ & $579$ & / & $620$ & ($93\%$) \\
\hphantom{$0$}$6$ & $371$ & $\times 10^{3}$ & $284$ & / & $306$ & ($93\%$) \\
\hphantom{$0$}$7$ & $114$ & $\times 10^{5}$ & $204$ & / & $212$ & ($96\%$) \\
\hphantom{$0$}$8$ & $410$ & $\times 10^{6}$ & $93$ & / & $94$ & ($99\%$) \\
\hphantom{$0$}$9$ & $170$ & $\times 10^{8}$ & $95$ & / & $97$ & ($98\%$) \\
$10$ & $794$ & $\times 10^{9}$ & $57$ & / & $60$ & ($95\%$) \\
\bottomrule

    \end{tabular}
    \caption{The number of senses of a word; the number of possible annotations per word; the number of annotated words out of the total in WordNet}
    \label{tab:annotationcollected}
\end{table}

\subsection{Qualitative Findings} \label{sec:meetings}

We held individual meetings with annotators at the end of each annotation phase to discuss their annotation.
The most common source for disagreement was the choice of prototypical sense(s), which could lead to radically different overall annotations. 
These disagreements usually resulted when WordNet had senses that were so similar that choosing which to make prototypical was arbitrary. 

Another common source of disagreement was idiolectical variation. Consider \word{host}, which refers to a person (\sense{host}{1}) or Communion wafer (\sense{host}{9}):
    
\noindent \begin{footnotesize}
\hspace{-.42em}\begin{tabular}{c p{6.4cm}} \tabspace
\sense{host}{1} & a person who invites guests to a social event (such as a party in his or her own home) and who is responsible for them while they are there \\ \tabspace
\sense{host}{9} & a technical name for the bread used in the service of Mass or Holy Communion 
\\ \tabspace
\end{tabular}
\end{footnotesize}
Two of the annotators labelled both of these senses as prototypical: to them, these senses were homonymous. However, the third annotator had been told at Catholic school that Communion bread `hosted' the Holy Spirit: to them, \sense{host}{9} was a metaphorical extension of \sense{host}{1}. This alternative interpretation was clearly represented in the feature transformation the annotator recorded.

\subsection{Agreement Measures and Results}

ChainNet data is complex, and not amenable to standard agreement measures. We use a multi-faceted approach.\footnote{For our inter-annotator agreement study, we include the annotation of all three annotators, as well as gold-standard annotation produced by one of the authors. 
For both inter- and intra-annotator agreement, we excluded wordforms which annotators indicated they did not know.}

\paragraph{Homonymy Agreement} If a word has multiple prototypical senses, then its senses are implicitly partitioned into disjointed clusters. 
We can investigate the degree to which annotators agree on homonymy by investigating whether their partitions align. 
For each word we compute the Adjusted Rand Index \citep[ARI;][]{hubert1985comparing} between the partitionings of each pair of annotators, and take the average over all words and all pairs. 
ARI is chance-corrected: $0$ indicates a random partitioning, while $1$ indicates perfect agreement. 
For inter-annotator agreement, the ARI was $.84$, while for intra-annotator agreement it was $.93$.

\paragraph{Label Agreement} To investigate whether annotators agree on the labels assigned to senses, we compute the mean pairwise percentage agreement, as well as \citeauthor{fleiss1971measuring}' Kappa.\footnote{We give non-prototypical senses the label of their incoming edge (either metaphor or metonymy).}
Measures for categorical classification such as Kappa require independence between datapoints. 
This does not hold for our labels, but Kappa is routinely used in the dependency parsing literature where the same issue applies: we report Kappa to allow comparison to that literature. 
In addition to reporting agreement for all senses (``All''), we additionally report the agreement when we filtered the data to only include a word if the annotators agreed on the choice(s) of prototype (Agree Prototype; ``AP''), or alternatively to only include a sense if the annotators agreed on which other sense it was connected to (Agree Connections; ``AC''). 
Results are presented in \cref{tab:label_agreement}.

\paragraph{Connection Agreement} 
To investigate whether annotators agreed on the connections between senses, we first compute the overall percentage of unlabelled connections that the annotators agreed on. 
The direction of connections emerges from the choice of prototype: to avoid cascading errors, we treat connections as undirected.
This is therefore equivalent to the undirected unlabelled attachment score (UUAS; see \citealp{gomez2015undirected}).
Once again, we additionally compute results when the data is filtered to only include words for which the annotators agreed on the choice(s) of prototype (AP).
Finally, we compute the percentage agreement over labels and connections jointly (i.e.\ the undirected labelled attachment score, ULAS). 
Results are presented in \cref{tab:attachment_agreement}.

\begin{table}
\small
    \centering
    \begin{tabular}{llc@{ ~}c@{ ~}cc@{ ~}c@{ ~}c}
 \toprule
     &            & \multicolumn{3}{c}{Percentage} & \multicolumn{3}{c}{Kappa} \\
     &            & \textit{All} & \textit{AP} & \textit{AC} & \textit{All} & \textit{AP} & \textit{AC} \\ \midrule
\multirow{4}{*}{\rotatebox[origin=c]{90}{Inter}} 
                & Prot.\        & $82$ &      &      & $.61$ \\
                & Metap.\     & $77$ & $88$ & $89$ & $.51$ & $.78$ & $.79$  \\
                & Meton.\     & $80$ & $88$ & $89$ & $.51$ & $.72$ & $.74$  \\
                & Any         & $70$ & $88$ & $89$ & $.54$ & $.84$ & $.85$  \\ \midrule
\multirow{4}{*}{\rotatebox[origin=c]{90}{Intra}} 
        & Prot.\        & $88$ &      &      & $.75$ \\
        & Metap.\     & $86$ & $92$ & $94$ & $.69$ & $.83$ & $.87$  \\
        & Meton.\     & $87$ & $92$ & $94$ & $.64$ & $.78$ & $.81$  \\
        & Any         & $81$ & $92$ & $94$ & $.70$ & $.88$ & $.90$  \\ \bottomrule
    \end{tabular}
    \caption{Label Agreement}
    \label{tab:label_agreement}
\end{table}

\subsection{Agreement Analysis}\label{sec:agreement}

In our qualitative findings (\cref{sec:meetings}) we observed that differences in prototype selection could lead to radically different annotations, reducing agreement. 
This observation is borne out in our empirical results. 
When all datapoints are considered, the inter-annotator agreement for labels is ${\kappa=.54}$. 
However, when the data is filtered to only include words where the annotators agreed on the prototypes (AP), this rises to ${\kappa=.84}$.
This same finding is reflected in connection agreement (UUAS $75\%$ for all vs.\ $91\%$ for AP).
For homonymy the inter-annotator agreement is high (${\ARI=.84}$), suggesting that difficulties in prototype selection did not stem from issues in determining whether or not two senses were related, but rather issues in choosing the best prototype from a group of closely related senses.

When we only include senses where the annotators agreed on connections (AC), label agreement is highest (${\kappa=.85}$).
This is especially true for metaphor labels, where agreement improves from~${\kappa=.51}$ to~${.79}$ under AC.
This gives credence to our operationalisation of metaphor as a relation: it is difficult to say whether a meaning of a word is metaphorical without knowing which other meaning it is being compared to, but agreement is high when annotators compare the same meanings.

Knowing the prototypical sense is nearly as good as knowing all the connections (${\kappa=.84}$ for AP vs.\ $.85$ for AC). 
One explanation for the similarity of these results could be that knowing the prototype(s) is sufficient to recover the connections.
This explanation is supported by the connection agreement, which improved from $75\%$ UUAS to $91\%$ in AP: when annotators agreed on the prototypical sense(s), they  agreed on the connections.

For the intra-annotator agreement study, the results pattern the same as for inter-annotator agreement, but are higher across the board 
(e.g.\ ${\kappa=.70}$ for unfiltered label agreement compared to $.54$ for inter). 
This is expected, as this setting will negate the effects of idiolectical variation. However, the ceiling for intra-annotator agreement remains limited by the sense inventory we used.

\begin{table}
\small
    \centering
    \begin{tabular}{lcccc} \toprule
    & \multicolumn{2}{c}{UUAS} & \multicolumn{2}{c}{ULAS} \\ 
    & \textit{All} & \textit{AP} & \textit{All} & \textit{AP} \\ \midrule
Inter & $75$ & $91$ & $65$ & $82$ \\
Intra & $82$ & $93$ & $76$ & $88$ \\
\bottomrule
    \end{tabular}
    \caption{Connection Agreement}
    \label{tab:attachment_agreement}
\end{table}

It is difficult to analyse the agreement of feature transformations because they are encoded in natural language.
As an informal experiment, we manually aligned the feature transformations for every pair of senses that two of the annotators agreed exhibited metaphor.
The result was positive ($93\%$ of metaphors had some form of overlap in their feature transformations), but we see this result as indicative rather than definitive.
Agreement values for virtual senses and split senses are also difficult to estimate, since these were very low frequency occurrences: there are only $19$ split and $258$ virtual senses in the entire dataset.

\section{Polysemy Parsing} \label{sec:baseline}

When presenting a new dataset, it is common to also present a baseline model that attempts to reproduce the collected data.
We call this task \defn{polysemy parsing}. 
For a word~$w$, the goal of polysemy parsing is to produce a parse of its senses~$\calS_w$, where a parse consists of a label for each sense (prototype, metaphor, or metonym), and a connection for each metaphor and metonym. 
Slippages, virtual senses, and split senses are not treated.

Polysemy parsing models do not have an obvious general application, but they could be used to synthesise annotations for the rest of the English WordNet, or to generate ChainNets for other languages. 
To investigate this possibility, we implement two polysemy parsing baselines.

\subsection{Two Baseline Models}

\paragraph{MPD+MST} 
Our first baseline predicts labels using a metaphorical polysemy detection \citep[MPD;][]{maudslay-teufel-2022-metaphorical} model, and then predicts connections by finding the minimum spanning tree (MST).
For each sense ${s\in\calS_w}$ we retrieve a $k$-dimensional sense-embedding ${\ve_s \in \real^k}$ \citep[from][]{scarlini2020sensembert}, and pass this through a linear layer and softmax to get probabilities for the three labels. 
Whichever sense has the highest prototype probability is taken as prototypical; the rest are given whichever label has the highest probability.\footnote{This ensures that at least one sense is a prototype.}
We compute the MST based on the distances between senses in the embedding space.

\paragraph{Biaffine} 
For a second baseline, we adapt a graph-based dependency parsing model, the biaffine parser \citep{dozat2017deep}. 
In dependency parsing, the goal is to find dependencies between words in a sentence. 
Graph-based parsers learn a function that predicts a score for each possible directed edge between a pair of words; a parse is then extracted by finding the maximum spanning tree over those scores using the \citetalias{chu1965shortest}/\citeauthor{edmonds1967optimum}'s algorithm \citep{mcdonald-etal-2005-non}.
To score an edge from a head sense~${h\in\calS_w}$ to a dependent sense~${d\in\calS_w}$, our adapted biaffine parser first retrieves the sense embeddings of those senses,~$\ve_h$ and~$\ve_d$. 
Using a pair of multi-layer perceptrons (MLPs), we process these into $k'$-dimensional head and dependent representations $\vh$ and $\vd$ respectively:\footnote{Because our input is a set of senses (not a list of words), we skip the contextualisation step of the standard biaffine parser, which is typically achieved using an LSTM \citep{hochreiter1997lstm}.}
 \begin{align}
 	\vh = \MLP_\text{head}(\ve_h) && \vd = \MLP_\text{dep}(\ve_d)
 \end{align}
The score is then computed using a biaffine function, which consists of a matrix $U$ ($\real^{k'} \times \real^{k'}$), a vector $\vw$ ($\real^{2{k'}}$), and a scalar $b$ ($\real$):
\begin{align}
\biaff(\vx, \vy) &= \vx^\top U \vy + \vw^\top (\vx \circ \vy) + b \\ 
\score(p \rightarrow s) &= \biaff(\vh, \vd)
\end{align}
Because the output for polysemy parsing is a forest (not a tree), we add an additional `root' sense to the input, and treat prototypical senses as being connected to this root. 
We embed the root as the mean of the sense embeddings of a word. 
A second biaffine function with its own parameters is used to predict edge labels. 
This need only distinguish between metaphor or metonymy, since prototypical senses are implicitly those connected to the root.

\begin{table}
\small
    \centering
    \begin{tabular}{llccc}
\toprule
 & Model & LOS & UUAS & ULAS \\ \midrule
\multirow{3}{*}{\rotatebox[origin=c]{90}{$1$ best}} & Random & $35$ & $41$ & $28$ \\
& MPD+MST & $\bm{51}$ & $52$ & $\bm{43}$ \\
& Biaffine & $50$ & $\bm{57}$ & $\bm{43}$ \\ \midrule
\multirow{3}{*}{\rotatebox[origin=c]{90}{$n$ best}} & Random & $42$ & $53$ & $36$ \\
& MPD+MST & $63$ & $68$ & $55$ \\
& Biaffine & $\bm{65}$ & $\bm{71}$ & $\bm{57}$ \\
\bottomrule
    \end{tabular}
    \caption{Baseline Results}
    \label{tab:baseline_results}
\end{table}

\subsection{Experimental Setup}

For evaluation metrics, we use labelling accuracy (label-only score, LOS), as well as UUAS and ULAS. 
In addition to evaluating the top prediction of each model (``$1$ best''), we additionally compare models when $n$ alternative high-scoring parses are considered (where $n$ is the number of senses of a word; ``$n$ best'').\footnote{To find each alternative parse, we removed an edge from the predicted parse, and recomputed the MST without that edge, then evaluated whichever parse from the options scored the highest under UUAS.} 
We do this to attempt to control against idiolectical variation in the gold standard which is used as evaluation data.
Significance values are computed using a two-tailed Monte Carlo permutation test ($r{=}10{,}000$; $\alpha{=}0.01$), with \citeauthor{bonferroni1936teoria} correction. Further details are in \cref{app:baseline}.

\subsection{Results} \label{sec:baseresults}

Results are presented in \cref{tab:baseline_results}. 
Both baselines perform significantly better than random in all settings. 
For the single-output comparison, under LOS and ULAS the two baselines were statistically indistinguishable, but for UUAS, the biaffine baseline performed significantly better than MPD+MST ($57$ compared to $52$). 
The lower performance of the biaffine parser under LOS and ULAS reflects a weakness in its labelling component, which is trained separately. 

When considering $n$ alternative predictions, every result is significantly higher than it was previously. 
In this setting, the MPD+MST and biaffine baselines are again significantly better than random, and the margin of their advantage is increased. 
For instance, the MPD+MST baseline gains $12$ points on LOS, while a random baseline improves by $7$.
This is most pronounced for the biaffine parser, which attains $71$ points UUAS when $n$ alternative parses are considered, up from $57$ in the single-output setting.
While the biaffine parser had difficulties predicting a single parse, it performed better when predicting several likely parses.

The performance of the biaffine parser may have been limited because it was trained using the out-of-the-box hyperparameters recommended by \citet{dozat2017deep}, or because of the alterations we made to adapt it for our task, specifically the lack of a contextualisation step and the way we compute the root embedding.
Nevertheless, these results represent a proof-of-concept, and suggest that future work may be able to synthesise ChainNet annotations for the rest of WordNet. 

\section{Reflections on Chaining} \label{sec:discussion}

A side effect of ChainNet's construction is that we have brought the theory of chaining \citep{lakoff1987women} into contact with a large amount of real-world data. 
In this section, we comment on two  points of interest that have arisen through this process.

\paragraph{Chains Evidence Multistability} 
A word's senses can be connected into chains in multiple valid ways that are sometimes radically different.
This is analogous to syntactic ambiguity (e.g.\ ``we saw her duck''), except that for chaining these situations are the rule, rather than the exception.
Evidence for this comes from multiple sources.
In our inter-annotator agreement study (\cref{sec:agreement}), we found that differences in the choice of prototypical sense(s) reduced agreement. 
In our qualitative analysis (\cref{sec:meetings}), we found that disagreements were sometimes caused by idiolectical variation between annotators (e.g.\ for the word \word{host}), and that the different annotation versions that resulted were usually internally-consistent.
In our evaluation of polysemy parsing baselines (\cref{sec:baseline}), we found that the performance of the biaffine parser improved substantially in a top-$n$ decode setting: the model could identify multiple high-scoring parses, but had difficulties in selecting a single ``best'' parse.
To address the effects of multistability, future ChainNet iterations could provide multiple valid parses for each word.
Alternatively, chains could be operationalised as cyclic graphs rather than trees, in which a nexus of senses serves as the prototype as opposed to a single sense.

\paragraph{Cognitive Chains ${\bm{\neq}}$ Etymological Chains}
We told annotators to prioritise their perception of words, and to override historical reality if necessary (\cref{sec:core}). 
For an indication of the effect this had, we can compare homonymy in ChainNet to etymological data. 
More specifically, we can retrieve clusters of interconnected senses from ChainNet, and compare these to clusters that are based on etymological derivation \citep[from][]{maudslay-teufel-2022-homonymy}. 
For $17\%$ of words, ChainNet's clusters are different from etymological clusters.
In a large majority of these cases ($95\%$), words exhibit homonymy in ChainNet where they do not in the etymological data.
That is to say, ChainNet's homonyms are more fine-grained than etymological homonyms: the mean number of clusters per word in ChainNet is $1.23$, but in etymological data it is $1.03$. 
One possible cause for this could be chaining.
Suppose that a word has a chain of senses that have emerged over time.
If an intermediate sense falls out of use, the senses of the word could form two disconnected chain fragments. 
This situation is illustrated below:
\begin{center}
    \includegraphics[trim={0cm .12cm 0cm 0cm},clip, width=0.75\columnwidth]{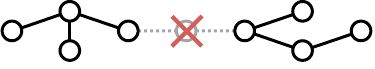}
\end{center}
The result is cognitive homonymy, without etymological homonymy.
A real-world example of this is \word{bridge}: the sense \sense{bridge}{9} (a deck of a ship) is derived from \sense{bridge}{1} (a structure to cross an obstacle), but this relation is no longer commonly perceived (see \cref{sec:core}).
We argue that future lexical-semantic resources should clearly distinguish between cognitive and etymological versions of phenomena, since the two do not always align.

\section{Conclusion}

We present ChainNet, a new dataset of metaphor and metonymy.
Rather than identifying instances of these phenomena in text, ChainNet instead reveals the sub-surface complexity of senses in WordNet. 
In addition to identifying inter-sense relations, ChainNet also provides a natural language explanation for each metaphor, in the form of a feature transformation.
This information can be used in tandem with other grounded information from WordNet to study metaphor and metonymy meaning.

ChainNet opens up new modelling possibilities.
Multimodal data from WordNet could be used to study metaphor meaning.
One direction would be to build cognitive models of metaphor generation, for example by adapting the task formulation of \citet{sun-etal-2021-computational}. 
Another direction would be to investigate whether ChainNet evidences patterns of conceptual metaphor or logical metonymy, for example using clustering algorithms. 
Future annotation work could extend ChainNet to other languages, or could refine ChainNet with additional relation types.

\section*{Data Availability Statement}

ChainNet and other accompanying material are freely available at \url{https://github.com/rowanhm/ChainNet}.

\section*{Ethics Statement}

The annotation collection component of this work involved working with participants; for this, we obtained approval from the Ethics Committee of the Department of Computer Science and Technology at the University of Cambridge. 
Annotators were paid £14.72 per hour of annotation work.

\section*{Acknowledgements}

The first author would like to thank the three students who carried out the annotation work, although ethical approval requires they remain anonymous. 
He would also like to thank Magdalene College, who supported him for the duration of this project, and the Alan Turing Institute and Isaac Newton Trust, who together provided the funding that enabled data collection.

\section*{Bibliographical References}
\bibliographystyle{lrec-coling2024-natbib}
\vspace{-2em}
\bibliography{lrec-coling2024-example, anthology}

\section*{Language Resource References}
\bibliographystylelanguageresource{lrec-coling2024-natbib}
\vspace{-2em}
\bibliographylanguageresource{languageresource}

\begin{figure*}
    \centering
    \includegraphics[trim={0cm 4.1cm 0cm 0cm},clip,width=\textwidth]{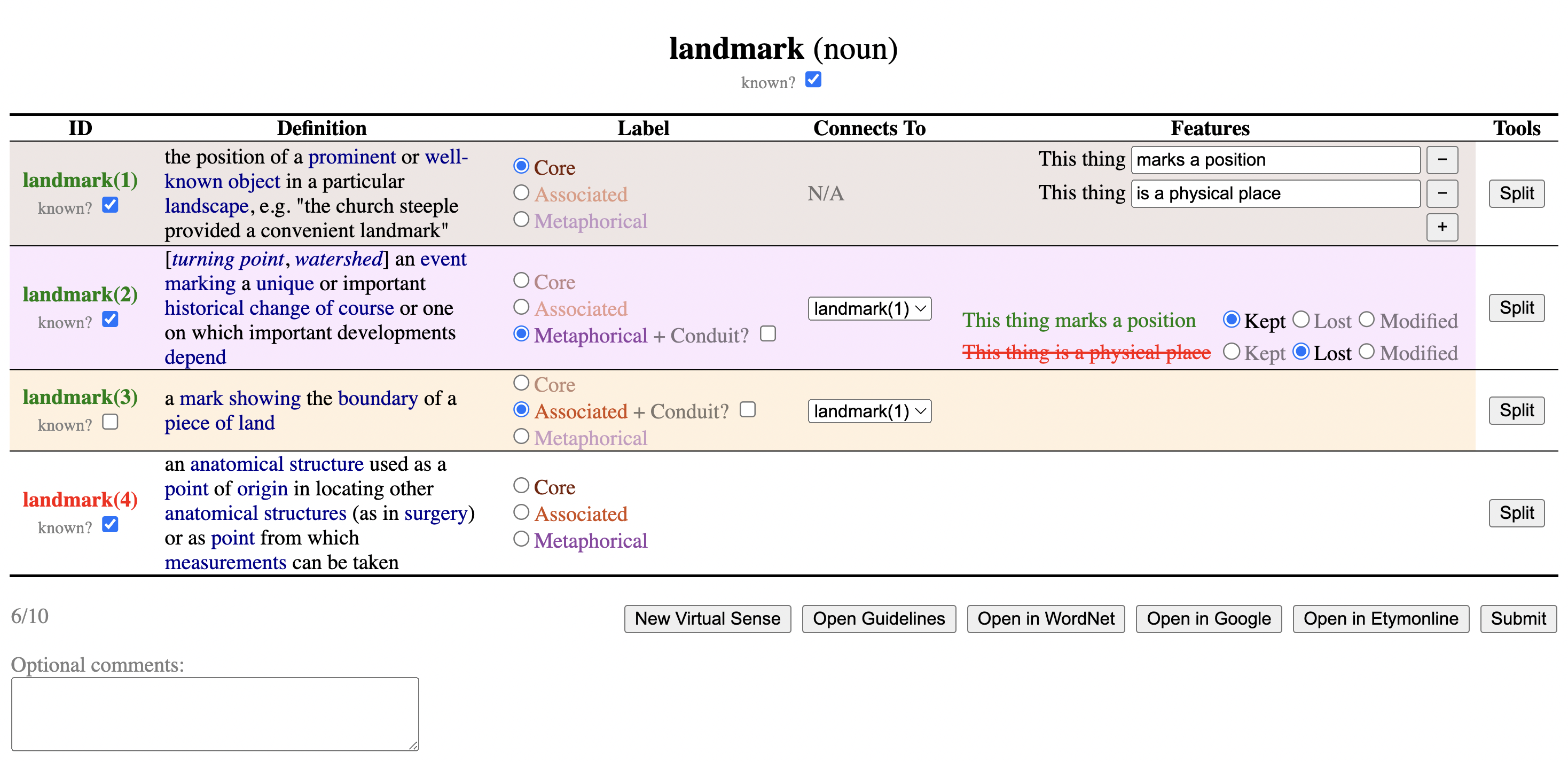}
    \caption{Screen capture of the annotation tool, with incomplete annotation for \word{landmark}}
    \label{fig:gui}
\end{figure*}

\appendix

\section{Calculating the Number of Possible Annotations}

Here we show how we compute the number of possible annotations of a word~$w$ that has $n$ senses (shown in \cref{tab:annotationcollected}), excluding features, split senses, and virtual senses. We denote this quantity $a_n$. 
This quantity corresponds to the number of possible forests of rooted trees with labelled edges. 

Assuming that only one sense can be prototypical (i.e.\ ignoring homonymy), we can take the number of possible annotations as:
\begin{align}
    a_n^1 = \underbrace{n^{n-2}}_{\text{(1)}} \, \times \, \underbrace{{k}^{n-1}}_{\text{(2)}} \, \times  \, \underbrace{n}_{\mathclap{\text{(3)}}} 
\end{align}
where the terms correspond to:
\begin{enumerate}[label={(\arabic*)}]
    \item The number of possible undirected trees with $n$ elements, given by \citeauthor{cayley1889trees}'s formula.
    \item The number of possible edge labellings for a tree with $n$ elements (and hence ${n-1}$ edges), and $k$~possible edge labels (in our case~${k=2}$, for metaphor and metonymy).
    \item The number of possible roots of a tree of size~$n$ (each element could be the prototype).
\end{enumerate}
We need to adapt this to accommodate homonymy (multiple prototypes). Let $[n]$ denote a set with $n$ elements, containing all the natural numbers up to and including $n$, i.e.\ ${[n]=\{1, 2, ..., n\}}$. 
These numbers correspond to the indices of the senses of $w$.
Furthermore, let $B_s$ be a set containing all possible partitions of a set $s$. 
For example, for the set ${[2]=\{1, 2\}}$, ${B_{[2]} = \{ \{ \{1\}, \{2\} \}, \{ \{1, 2\} \} \}}$. 
Each partition corresponds to a particular way of splitting the senses of a word into disjoint sets, i.e.\ homonyms. 
We use this to calculate $a_n$:

\begin{align}
    a_n &= \sum_{\calP \in B_{[n]}} \prod_{\calS \in \calP} a_{\setsize{\calS}}^1
\end{align}
The inner term gives us the total number of possible annotations for a particular partitioning. The outer summation covers all possible partitionings. 

\section{The Annotation Interface} \label{app:interface}

The annotation for a word is entered by completing a table, where each row of the table contains one of the word's senses. 
This interface is shown in \cref{fig:gui}.  
At the bottom of the table, there are buttons which can be used to add virtual senses, to open the annotation guidelines, to search for the word in various other resources, and finally to submit the annotation.
Annotation can only be submitted if it is complete, otherwise the system will not accept the submission.
From left to right, the columns are:

\paragraph{ID} 
This column contains the ID of each sense, as well as a check box which annotators can use to mark whether they know the sense (by default it is checked). 
Whole words can be labelled as unknown by deselecting the check box below the word at the top of the screen; annotators are required to annotate every word and sense, including those they did not previously know.
A red sense ID indicates that a sense has incomplete annotation, whereas green indicates completion.

\paragraph{Definition} 
This column contains the WordNet definition for each sense. 
Sometimes these definitions are vague. If this is the case, annotators can hover their mouse over blue words in the definition.
This causes a popup bubble to appear which reveals the definition of the word in question.

\paragraph{Label}
Annotators use radio buttons to decide which label to apply to each sense.
If a sense is labelled as a metaphor or a metonym, the option appears to mark the sense as a conduit. 
Conduits can be extended by any other sense (see \cref{sec:guidelines}).

\paragraph{Connects To}
If the annotator labels a sense as a metaphor or metonym, then a drop-down list will appear in this column.
Annotators use this list to choose which sense is being extended.
Options that are invalid (such as the same sense that is being labelled) are greyed out.

\paragraph{Features}
If a sense is extended by a metaphor (e.g.\ \sense{landmark}{1} in \cref{fig:gui}), then a button appears in this column to add features.
Clicking the button causes a feature to appear.
Each feature has an empty text box that can be filled in.
Meanwhile, if a sense is a metaphor (\sense{landmark}{2} in \cref{fig:gui}), this column will contain a copy of each feature of the sense it extends. 
These copies are each accompanied by radio buttons to indicate whether the feature is kept, lost, or modified. 
If a feature is marked as modified then a text box appears containing the same text as the feature. 
This text box is used to record the modification.

\paragraph{Tools}
This column contains buttons to perform additional operations.
For most senses the button gives annotators the option to split the sense.
If clicked, that row of the column will split in two, the definitions will become editable, and the button will change to give annotators the option to re-merge the senses.
If a virtual sense is added, a new row is added to the table; in that case, the button gives annotators the option to delete the virtual sense.

\section{Baseline Training Details} \label{app:baseline}

\paragraph{Training Data} 
Polysemy parsing does not treat virtual senses or split senses.
To train models, we process ChainNet to remove these features.
More specifically, we remove each split sense by re-merging it and giving the combined sense whichever annotation belonged to its non-metaphorical component. 
We remove each virtual sense by connecting all of its children to its parent.
We then randomly divide the resultant dataset into train, development, and test partitions (an ${80{:}10{:}10}$ split of the $6500$ words in ChainNet).

\paragraph{Training Procedure}
We implement models in PyTorch \citep{adam2019pytorch}, and optimise the cross entropy loss in each case using AdamW \citep{loshchilov2017adamw}.
Every epoch, we compute the loss on the development set. 
When the loss fails to decline for $8$ epochs we recover the best model.
The first time this happens we divide the learning rate by $10$ and continue training; the second time we terminate model training.

\paragraph{Hyperparameters} 
We trained in batches of $32$ words, with a learning rate of ${\alpha={5{\times}10^{-5}}}$, and $\beta_1$ and $\beta_2$ for AdamW both set to $0.9$.
For the biaffine parser the dropout \citep{srivastava2014dropout} was $0.33$; $k'$ was $100$ for the label predictor and $2048$ for the edge predictor (to match the dimensionality of the embeddings, ${k=2048}$ from \citealp{scarlini2020sensembert}). 
These parameters were chosen following \citet{dozat2017deep}.

\end{document}